\documentclass[runningheads]{llncs}
\usepackage[T1]{fontenc}
\usepackage{graphicx}
\usepackage{amsmath}
\usepackage{caption} 
\usepackage{hyperref}

\begin{document}
%

\title{EndoDepth: A Benchmark for Assessing Robustness in
Endoscopic Depth Prediction}

%

\titlerunning{EndoDepth Benchmark}
%


\author{ Ivan Reyes-Amezcua\inst{1} \and
Ricardo Espinosa \inst{2,4} \and
Christian Daul \inst{4}  \and
Gilberto Ochoa-Ruiz\inst{3} \and 
Andres Mendez-Vazquez \inst{1} }
\authorrunning{I. Reyes et al.}
%
\institute{CINVESTAV Unidad Guadalajara, Mexico \and Universidad Panamericana, Facultad de Ingenieria, Aguascalientes, Mexico \and
Esc. de Ingenieria y Ciencias, Tecnologico de Monterrey, Monterrey, N.L., Mexico  \and
CRAN (UMR 7039), Université de Lorraine and CNRS, Nancy, France
}

\maketitle              
\vspace{-7mm}

\begin{abstract}
Accurate depth estimation in endoscopy is vital for successfully implementing computer vision pipelines for various medical procedures and CAD tools. In this paper, we present the EndoDepth benchmark, an evaluation framework designed to assess the robustness of monocular depth prediction models in endoscopic scenarios. Unlike traditional datasets, the EndoDepth benchmark incorporates common challenges encountered during endoscopic procedures. We present an evaluation approach that is consistent and specifically designed to evaluate the robustness performance of the model in endoscopic scenarios. Among these is a novel composite metric called the mean Depth Estimation Robustness Score (mDERS), which offers an in-depth evaluation of a model's accuracy against errors brought on by endoscopic image corruptions. Moreover, we present SCARED-C, a new dataset designed specifically to assess endoscopy robustness. Through extensive experimentation, we evaluate state-of-the-art depth prediction architectures on the EndoDepth benchmark, revealing their strengths and weaknesses in handling endoscopic challenging imaging artifacts. Our results demonstrate the importance of specialized techniques for accurate depth estimation in endoscopy and provide valuable insights for future research directions. 
\hfill \break

\centering{\url{https://github.com/Ivanrs297/endoscopycorruptions}}
\end{abstract}

\section{Introduction}
\label{sec:introduction}



Endoscopic procedures offer crucial insights into hollow organ interiors, providing essential tissue information. However, challenges in controlling the endoscope's trajectory and viewpoint complicate examinations, leading to suboptimal views and incomplete scans, hampering lesion diagnosis. Traditional multi-view stereo methods like structure-from-motion (SfM) \cite{leonard2018evaluation}, shape-from-shading (SfS) \cite{Ren_2017}, SLAM \cite{ma2019real} excel in natural light settings but face limitations in endoscopic scenarios due to sparse textures and sudden illumination changes.

On the other hand, recent strides in depth estimation have relied on high-quality training data, either annotated or not. For instance, supervised learning-based monocular depth estimation methods have gained a lot of traction on conventional imaging, leveraging recent strides in deep learning architectures and large amounts of data acquired using other modalities to train models to predict depth from single images effectively. However, obtaining ground truth data for endoscopic imaging poses challenges, hindering the application of supervised methods \cite{EigenPF14,8010878,liu2019dense}.

\begin{figure}[!b]
    \centering
    \includegraphics[scale=0.18]{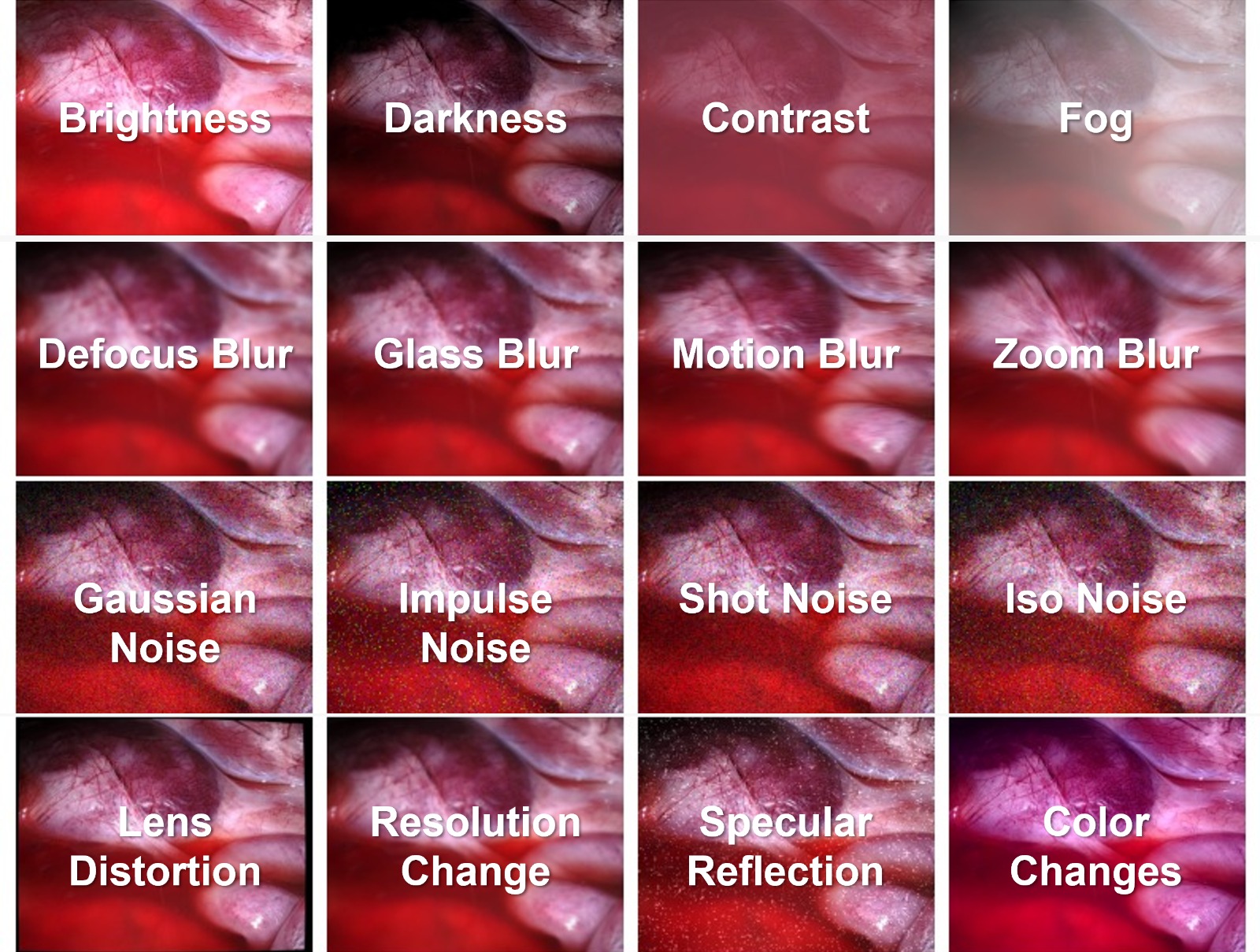}
    \caption{Example of the EndoDepth Corruptions on an image from the SCARED-C dataset. In the last row are image corruptions that significantly affect the performance of monocular depth estimation in endoscopic imaging, including lens distortion, resolution alterations, specular reflection, and color changes.}
    \label{Fig1}
\end{figure}

To address the lack of ground truth for depth estimation, several studies have explored unsupervised techniques using auxiliary supervisory signals from data. Many leverage view synthesis based on structure-from-motion principles for self-supervision, using predicted depth and pose to reconstruct target frames from source frames. The discrepancy between re-projected and source frames serves as the learning objective. However, these methods face challenges in endoscopic imaging due to extreme acquisition conditions and brightness constancy violations\cite{bian2019unsupervised,godard2019digging}.

Nevertheless, when it comes to extreme acquisition conditions like those encountered in endoscopy, the aforementioned methods face unique challenges. In endoscopy, scene illumination is heavily influenced by the orientation of the endoscope relative to the tissue surface. The captured images frequently suffer from under or over-exposure, depending on the surface shape, and are susceptible to specular reflections. Additionally, they may depict textureless scenes and become foggy due to rapid increases in endoscope temperature within human cavities, etc.

Despite advancements in supervised and self-supervised depth prediction models in endoscopy, there remains a significant gap in understanding their robustness to out-of-distribution challenges, particularly in dealing with endoscopy-specific corruptions like adverse exposure and sensor malfunctions \cite{ma2019real}. Conventional machine learning-based visual perception models are often sensitive to subtle changes in lighting, noise, and texture variations, leading to compromised depth prediction accuracy \cite{OZYORUK2021102058,shao2022self}. While progress has been made with datasets like EndoSfMLearner, SCARED, and Hamlyn \cite{OZYORUK2021102058,allan2021stereo,recasens2021endo}, a crucial gap remains: the absence of a robustness benchmark tailored specifically for developing resilient and scalable endoscopy depth prediction systems.

To address this gap, our work introduces the first steps toward creating robust and dependable endoscopy depth prediction systems by introducing a base benchmark on the SCARED dataset. Our benchmark meticulously simulates common corruptions inherent to endoscopy environments. As illustrated in Figure \ref{Fig1}, we categorize sixteen types of corruptions into four main groups: i) lighting conditions, ii) data processing complications, iii) sensor malfunctions and movements, and iV) endoscopy common corruptions. These corruptions, further stratified by severity, encompass a wide range of scenarios that induce image distortions, texture alterations, or degraded visual quality. The code for this paper is available on GitHub repository at \url{https://github.com/Ivanrs297/endoscopycorruptions}

The main contributions of this paper, based on the new dataset and robustness evaluation frameworks as summarized as follows:

1) We introduce EndoDepth Benchmark, the first systematically designed robustness evaluation for Supervised and Self-Supervised depth estimation models, capable of handling data corruptions, sensor failures, and style shifts.

2) We propose a new composite metric called mean Depth Estimation Robustness Score (mDERS) by capturing both accuracy and error resistance in the context of endoscopic image corruptions.

3) We evaluate the robustness of 4 state-of-the-art self-supervised depth prediction models in endoscopy using our novel dataset called SCARED-C

4) Drawing from our findings, we provide an in-depth discussion and analysis of the key design considerations for developing more resilient Self-supervised depth prediction models suitable for reliable, scalable, and practical applications.\newline


\vspace{-7mm}

\section{Related Work}
\label{sec:related_work}

Several studies have demonstrated that Deep Neural Networks (DNNs) are susceptible to common corruptions, such as blur, Gaussian noise, and translations \cite{hendrycks2018benchmarking}. In the field of computer vision for endoscopy, several works \cite{Endo4IE,OZYORUK2021102058,shao2022self} have shown that the performance of models for depth estimation can significantly decline when faced with artifacts and other challenging imaging conditions unique to endoscopy. For instance, acquisition perturbations due to blurring, and specular reflections due to movement of the point of the endoscope are significantly stronger than in natural imagery scenarios. To systematically investigate the robustness of DNN against image corruption, different benchmarks have been introduced. These benchmarks were initially developed in the domain of image recognition \cite{hendrycks2018benchmarking} and have since been extended to various tasks including object detection and semantic segmentation \cite{LIU2023175}. To address robustness in depth estimation for autonomous driving and real-world applications, a recent work named RoboDepth \cite{kong2023robodepth} has been introduced. By systematically evaluating depth estimation models under varied conditions, RoboDepth provides insights into the models' robustness and performance in challenging real-world scenarios.

However, the endoscopic domain presents unique challenges that require specific attention and evaluation. Given the distinct characteristics and demands of endoscopic scenarios, a dedicated benchmark tailored to this domain becomes imperative. Existing benchmarks may not fully capture the intricacies and challenges inherent in endoscopic imaging, such as limited field of view, motion artifacts, specular reflections, and variations in tissue appearance.

In previous years, the most commonly used metric for assessing the robustness of depth prediction in general tasks has been the Mean Corruption Error (mCE) \cite{hendrycks2018benchmarking}. This metric requires a base model to evaluate the performance of other models across various corruptions \cite{kong2023robodepth}. However, using a reference model can bias the mCE results of other models, making them dependent on the chosen baseline model.

Our proposal incorporates the most common error metrics for depth prediction Abs Rel, Sq Rel, RMSE, RMSE log and accuracy measures a1, a2, a3, forming a metric that relies solely on evaluating models without needing base models. Additionally, our metric provides a single output that establishes a robust relationship between errors and accuracy, enabling a straightforward and fair evaluation of the robustness of depth prediction models for endoscopy.




\section{The EndoDepth Benchmark Design}

This section explores the robustness study we carried out as part of our quantitative analysis, emphasizing the clinical applications of these models' implications and their resistance against common disturbances observed during endoscopic examinations. Our approach is inspired by the framework developed in \cite{hendrycks2018benchmarking} for classification tasks. In this seminal work, the \textit{mean Corruption Error} (mCE) was proposed to function as a standardized aggregate performance metric that intended to serve as a proxy of how robust a model is to different kinds of corruption in image classification tasks.

To assess the robustness of endoscopic depth estimation methods, we propose a metric that encapsulates the model's performance under a variety of realistic and challenging endoscopic procedure-specific conditions. Specifically, we introduce a metric for monocular depth estimation in endoscopy, which we'll refer to as the \textit{mean Depth Estimation Robustness Score} (mDERS).

\vspace{-2mm}

\subsection{Mean Depth Estimation Robustness Score (mDERS)}

Accuracy and robustness are crucial in depth estimation, particularly where image degradations impact performance. Traditional metrics often fail to capture model robustness against real-world image corruptions. To address this, we introduce the mean Depth Estimation Robustness Score (mDERS), a composite metric designed to comprehensively assess the robustness of depth estimation models.

Given the error measures $e: \{Abs Rel, Sq Rel, RMSE, RMSE log\}$, these are metrics where lower values indicate better performance. Also, the accuracy measures $a: \{a1, a2, a3\}$, these are metrics where higher values indicate better performance. They quantify how accurately a model can estimate depth in various scenarios. We determine the mean of each error $\bar{e}$ and accuracy $\bar{a}$ metric across all severity levels for each corruption to combine these multi-level results into a single metric. This procedure of averaging reduces the impact of outliers and yields a reliable indicator of the model's performance in the face of various forms of corruption. Thus, the mean Depth Estimation Robustness Score (mDERS) is formulated as follows:

\vspace{-3mm}

\begin{equation}
    mDERS = \frac{\frac{1}{n} \sum_{i=1}^{n} \bar{a}_i}{1 + \frac{1}{m} \sum_{i=1}^{m} \bar{e}_i}  
\end{equation}

Where, the numerator \(\frac{1}{n} \sum_{i=1}^{n} \bar{a}_i\) calculates the mean accuracy across the \(n_a\) accuracy metrics. This part of the formula aggregates the accuracy measures by taking their sum and dividing them by the number of accuracy metrics (\(n_a\)), which is 3 in this context. Then, the denominator \(1 + \frac{1}{m} \sum_{i=1}^{m} \bar{e}_i\) accounts for the error metrics. It calculates the mean error across the \(m\) error metrics, which is 4 in this case. The addition of 1 ensures that the denominator is always greater than 1, preventing the overall score from becoming disproportionately high due to very low error values and maintaining a balanced scale.

Hence, by combining accuracy and error into a single, balanced score, mDERS essentially offers a comprehensive assessment of a depth estimating model's performance. A robust depth estimation is reflected in models with high mDERS values, which are both accurate and maintain low levels of estimate errors.

\subsection{Mean Corruption Error for Endoscopy Depth Estimation}

The notion of Mean Corruption Error (mCE) for endoscopy is presented in this subsection. In this context, mCE is a metric that measures a model's error response across different types and severity levels of corruption that are encountered during endoscopic procedures. We present a methodical procedure for computing mCE, which yields a normalized score indicating how robust the model is in comparison to a baseline or average performance. 

Let's define \(D_{f}^{s,p}\) as the depth estimation error of the model \(f\) under perturbation type \(p\) at severity level \(s\). Examples of appropriate error metrics to consider are the mean absolute error and the root mean squared error. The error metric selected can be customized to meet the needs of any given application. In our case, a wide variety of conditions found in endoscopic imaging negatively impact the quality of the acquired image and thus the depth estimate precision. Therefore, image degradations or artifacts found in endoscopy should be covered by use as possible perturbations to the depth estimate. The mCE for endoscopy depth estimation  is  formulated as follows:

\begin{equation}
    \text{mCE} = \frac{1}{C} \sum_{c=1}^{C} \left( \frac{\sum_{s=1}^{S} D_{f}^{s,c}}{\sum_{s=1}^{S} D_{\text{baseline}}^{s,c}} \right)
\end{equation}

\noindent where \(D_{f}^{s,c}\) is the depth estimation error of the model \(f\) for corruption type \(c\) at severity level \(s\). On the other hand,  \(D_{\text{baseline}}^{s,c}\) is the depth estimation error of a baseline model (or the average error across models) for the same perturbation type and severity level. This normalization step ensures that the scores are comparable across different perturbations. \(S\) is the number of severity levels considered for each perturbation type. \(C\) is the total number of perturbation types relevant to endoscopic depth estimation. Thus, the model's robustness to various endoscopic perturbations is summed up in a single score provided by the mCE. Better robustness is indicated by a lower mCE, which suggests that the model performs worse in demanding circumstances than the baseline. 
\\





Seven essential metrics are used to measure each model's performance under these circumstances: accuracy under threshold (a1, a2, a3), squared relative difference (sq\_rel), root mean square error (rmse), and root mean square error in logarithmic scale (rmse\_log). These measures offer a thorough understanding of the models' precision, accuracy, and general capacity to sustain performance integrity in challenging circumstances. For more details about these metrics please refer to \cite{godard2019digging}.

\section{Methods and Data}
\subsection{Data}
To investigate how the corruptions affect the performance of depth estimation, we conducted N perturbation experiments utilizing the SCARED dataset. The SCARED dataset is partitioned into training (15351 frames), validation (1705 frames), and test sets (551 frames). The SCARED dataset, introduced by Allan et al. \cite{allan2021stereo}. It consists of 35 endoscopic videos accompanied by point cloud and ego-motion ground truth annotations. 

Herein, we present a novel dataset, SCARED-C \cite{Ivanrs297_endoscopycorruptions}, which has been carefully selected to test the robustness of depth estimation models in endoscopic imaging settings. With 16 different types of corruption, SCARED-C covers a broad spectrum of typical endoscopic problems, each with five levels of severity. It is expected that the SCARED-C dataset will prove to be an invaluable asset for the medical community, enabling substantial advancements in the robustness and dependability of depth estimate algorithms. 

\begin{table}[b!]
\centering
\begin{tabular*}{\linewidth}{@{\extracolsep{\fill}} cccccc }
\hline
Metric & Monodepth2* & AF-SfMLearner & MonoViT & EndoSfMLearner \\ \hline
abs\_rel   & 100.00 & \textbf{77.60} & 92.04 & 133.39 \\
sq\_rel    & 100.00 & \textbf{66.75} & 101.30 & 143.73 \\
rmse       & 100.00 & \textbf{80.63} & 91.37 & 107.54 \\
rmse\_log  & 100.00 & \textbf{80.38} & 94.39 & 124.53 \\
a1         & 100.00 & 106.60 & 103.91 & \textbf{91.76}  \\
a2         & 100.00 & 100.32 & 98.99 & \textbf{98.00}  \\
a3         & 100.00 & 100.14 & 99.50 & \textbf{99.86}  \\ \hline 
\end{tabular*}
\caption{Mean Corruption Error (mCE) given in percentage (\%) of four models: Monodepth2, AF-SfMLearner, MonoViT, and EndoSfMLearner across a range of metrics. 
}
\label{table:mce_metrics}
\end{table}

\vspace{-2mm}

\subsection{The EndoDepth Benchmark Corruptions}

Image corruptions such as lens distortion, resolution changes, specular reflections, and color variations significantly affect monocular depth estimation in endoscopic imaging. These corruptions complicate spatial interpretation and depth perception. Addressing these issues is crucial for improving depth estimation models. We took some corruptions from \cite{hendrycks2018benchmarking} and provided a brief description of each relevant to endoscopic images. Robust modeling approaches are essential to mitigate these effects and ensure reliable depth measurements in endoscopic procedures.

\begin{itemize}
    \item \textit{Brightness}: Variations in illumination intensity across the image. It can be mathematically represented as $I' = \alpha I$, where $I'$ is the modified image, $I$ is the original image, and $\alpha > 1$ indicates increased brightness.
    \item \textit{Darkness}: A reduction in illumination intensity, opposite to brightness. It follows the same formula as brightness but with $\alpha < 1$.
    \item \textit{Contrast}: The difference in luminance or color that makes an object distinguishable. Mathematically, contrast modification can be represented as $I' = \alpha(I - \mu) + \mu$, where $\mu$ is the mean luminance of the image, and $\alpha$ controls the level of contrast.
    \item \textit{Fog}: Simulates the scattering of light due to particles in the air, leading to reduced visibility. It can be modeled as $I' = I(1 - \omega) + A\omega$, where $A$ is the atmospheric light, and $\omega$ represents the amount of fog.
    \item \textit{Defocus Blur}: Occurs when the image is out of the focal plane, resulting in a blur that is uniform across the image. It is often modeled by convolving the image with a disk-shaped kernel.
    \item \textit{Glass Blur}: Simulates the distortion caused by viewing through a rough glass surface. It is typically achieved by displacing pixels randomly in a manner that mimics the refraction through glass.
    \item \textit{Motion Blur}: Results from the rapid movement of either the camera or the subject, leading to streaking or blurring in the direction of motion. It can be modeled by convolving the image with a linear kernel aligned with the direction of motion.
    \item \textit{Zoom Blur}: Occurs during rapid zooming, causing radial streaks from the center of the image outward. It is modeled by radially blurring the image from a central point.
    \item \textit{Gaussian Noise}: Adds a statistical noise that follows a Gaussian distribution, represented as $I' = I + N(0, \sigma^2)$, where $N$ is the Gaussian distribution with mean 0 and variance $\sigma^2$.
    \item \textit{Impulse Noise}: Also known as salt-and-pepper noise, it randomly alters some of the image pixels to black or white, representing a sparse distribution of noise.
    \item \textit{Shot Noise}: Originates from the discrete nature of light itself, modeled as Poisson noise where the variance of the noise is signal-dependent.
    \item \textit{ISO Noise}: Graininess or noise introduced by high ISO settings in cameras, characterized by both luminance and color noise.
    \item \textit{Lens Distortion}: Causes straight lines to appear curved due to the optical design of lenses. It is often corrected by applying distortion correction algorithms.
    \item \textit{Resolution Change}: Involves altering the image's resolution, which can affect the apparent depth cues in an image. This can involve either downsampling or upsampling techniques.
    \item \textit{Specular Reflection}: Bright spots that occur when light directly reflects off shiny surfaces into the camera. These can create high-intensity regions that may mislead depth estimation.
    \item \textit{Color Changes}: Variations in color balance and saturation, which can be due to different lighting conditions or camera settings. This can affect the perceived depth and texture in the image.
\end{itemize}

\begin{figure}[t!]
    \centering
    \includegraphics[width=1\textwidth]{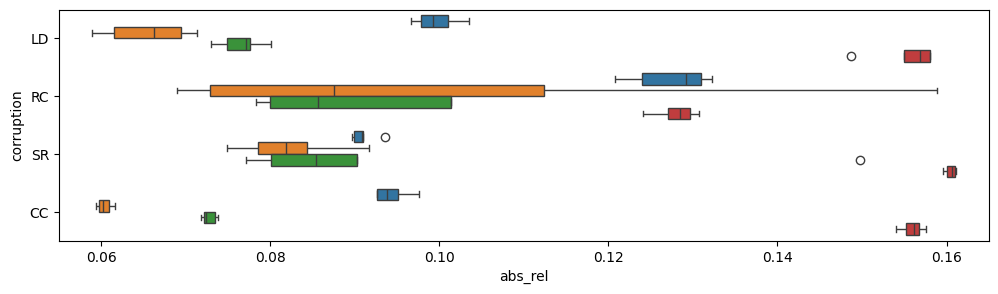}
    \caption{Boxplot for the absolute relative error of four depth estimation models: Monodepth2 (blue), AF-SfMLearner (orange), MonoViT (green), and EndoSfMLearner (red), under various image corruption types: Lens Distortion (LD), Resolution Change (RC), Specular Reflection (SR), and Color Changes (CC). }
    \label{boxplot}
\end{figure}

\vspace{-2mm}

\subsection{Evaluation of Robustness in Monocular Depth Prediction }

In this section, we provide an extensive assessment of the resilience of several state-of-the-art monocular depth prediction models using our mDERS. Considering the "Monodepth2" model broad use in the depth prediction community, our assessment uses it as a baseline. We use a set of 16 different image corruptions (see Figure 1), each at 5 different levels of severity, throughout the testing phase to systematically evaluate the robustness of the models (Figure \ref{boxplot}). For our experiments, we tested various recent state-of-the-art models for depth estimation and 3D reconstruction in endoscopy: AF-SfMLearner \cite{shao2022self}, MonoViT \cite{Zhao_2022}, and EndoSfMLearner \cite{OZYORUK2021102058}.


\begin{table}[b!]
\centering
\begin{tabular*}{\linewidth}{@{\extracolsep{\fill}} cccccc }

\hline
& Monodepth2 & AF-SfMLearner & MonoViT & EndoSfMLearner \\ \hline
mDERS & 0.2608 & \textbf{0.3134} & 0.2759 & 0.2332
\end{tabular*}
\caption{mDERS comparison between the four models. The model with the highest score, bolded, is the most robust.}
\label{table:mDERS}
\end{table}

\vspace{-2mm}

\section{Experiments and Evaluation}


 Table \ref{table:mce_metrics} summarizes our experiments, in which we the tested models with the well-known Monodepth2 model, which served as a baseline for calculating the mCE. Similarly, Figure \ref{boxplot} shows a visual representation of a model's robustness that shows the range and variability of the model's errors (abs\_rel).



\begin{figure}[t!]
    \centering
    \includegraphics[width=0.95\textwidth]{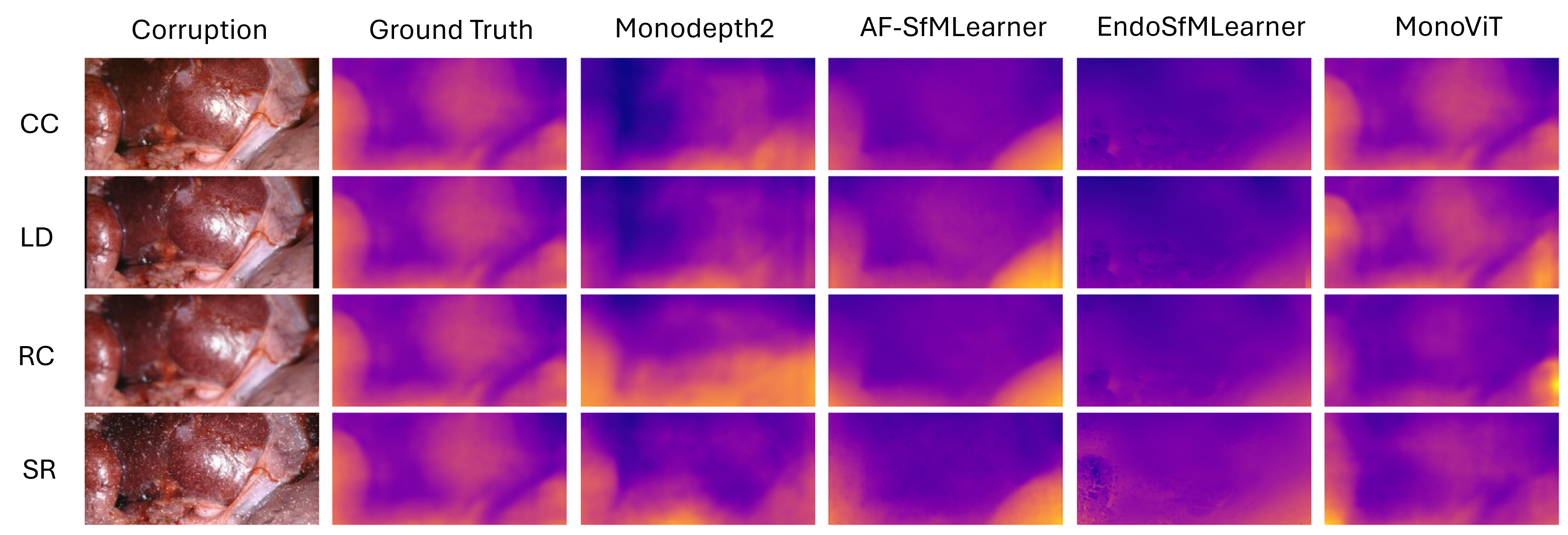}
    \caption{Visual comparison of depth estimation outputs from multiple models. The first column shows the endoscopic images using the endoscopy corruptions at a severity of 2 using our method, followed by ground truth image and the depth predictions from the Monodepth2, AF-SfMLearner, MonoViT, and EndoSfMLearner models, respectively. The corruptions used are: Lens Distortion (LD), Resolution Change (RC), Specular Reflection (SR), and Color Changes (CC).}
    \label{qualitative}
\end{figure}

A comparison of the Depth Estimation Robustness Score (mDERS) for the four state-of-the-art depth estimation models is shown quantitatively on the top of Figure \ref{qualitative}, which also shows a qualitative comparison of the estimated depth maps produced by the different models under various perturbations.

The model's overall performance in-depth estimation tasks is assessed using our score mDERS, which combines accuracy and error values. Our experiments reveal that AF-SfMLearner emerges as the most reliable model, boasting the highest mDERS value of 0.3134. This indicates its successful balance between depth prediction accuracy and low error rates. However, AF-SfMLearner shows susceptibility to Resolution Changes, a common perturbation in endoscopy procedures, as depicted in Fig. \ref{boxplot}. Conversely, EndoSfMLearner receives the lowest score with 0.232, suggesting a need for improvement in its robustness under the studied conditions, particularly concerning strong illumination changes, specular reflection and color changes, as illustrated in the box plots of Fig. \ref{boxplot}. 
The qualitative results in Fig. \ref{qualitative} are consistent with the findings of Fig. \ref{boxplot}: EndoSFMLearner under-performs under all types of corruptions, while AF-SFMLearner consistently produces good depth maps despite different types of artifacts.



\vspace{-4mm}

\section{Conclusion}

In this work, our  aim was to identify the weaknesses of the SOTA depth prediction model against imaging artifacts. By applying varying levels of severity across a range of corruptions, we can test the models' limits and pinpoint areas for improvement. The mDERS metric was introduced to evaluate depth prediction in endoscopy, showing a good performance in terms of modeling robustness. To foster further research in this area, the SCARED-C dataset is now publicly accessible on our GitHub repository \cite{Ivanrs297_endoscopycorruptions}.

\section*{Acknowledgments}
The authors wish to acknowledge the Mexican Council for Science and Technology (CONACYT) for their support in terms of postgraduate scholarships in this project, and the Data Science Hub at Tecnologico de Monterrey for their support on this project. 
This work has been supported by Azure Sponsorship credits granted by Microsoft's AI for Good Research Lab through the AI for Health program.

The project was also supported by the French-Mexican ANUIES CONAHCYT Ecos Nord grant 322537.



%
%
%
\bibliographystyle{splncs04}
\bibliography{main}
\end{document}